\documentclass[dvipsnames,nonacm, format=sigconf,screen = true, ,anonymous=false,review=false]{acmart}
\usepackage{mathtools} 
\usepackage{subcaption} 
\usepackage{booktabs}
\usepackage{todonotes}

\renewcommand{\epsilon}{\varepsilon}

\newcommand{\sobol}{Sobol\kern-0.15em\'{ } }

\AtBeginDocument{%
  \providecommand\BibTeX{{%
    \normalfont B\kern-0.5em{\scshape i\kern-0.25em b}\kern-0.8em\TeX}}}

\begin{document}

\title{Improving Algorithm-Selection and Performance-Prediction via Learning Discriminating Training Samples}

\author{Quentin Renau}

\affiliation{%
  \institution{Edinburgh Napier University}
  \city{Edinburgh}
  \country{Scotland, UK}}
  \email{q.renau@napier.ac.uk}

\author{Emma Hart}

\affiliation{%
  \institution{Edinburgh Napier University}
  \city{Edinburgh}
  \country{Scotland, UK}}
  \email{e.hart@napier.ac.uk}

\renewcommand{\shortauthors}{Q. Renau and E. Hart}

\begin{abstract}
The choice of input-data used to train algorithm-selection models is recognised as being a critical part of the model success. Recently, feature-free methods for algorithm-selection that use short trajectories obtained from running a solver as input have shown promise. However, it is unclear to what extent these trajectories reliably discriminate between solvers. We propose a \textit{meta} approach to generating discriminatory trajectories with respect to a portfolio of solvers. The algorithm-configuration tool \textit{irace} is used to tune the parameters of a simple Simulated Annealing algorithm (SA) to produce trajectories that maximise the performance metrics of ML models trained on this data. We show that when the trajectories obtained from the tuned SA algorithm are used in ML models for algorithm-selection and performance prediction, we obtain significantly improved performance metrics compared to models trained both on raw trajectory data and on exploratory landscape features.
\end{abstract}


\begin{teaserfigure}
\centering
  \includegraphics[width=0.7\linewidth]{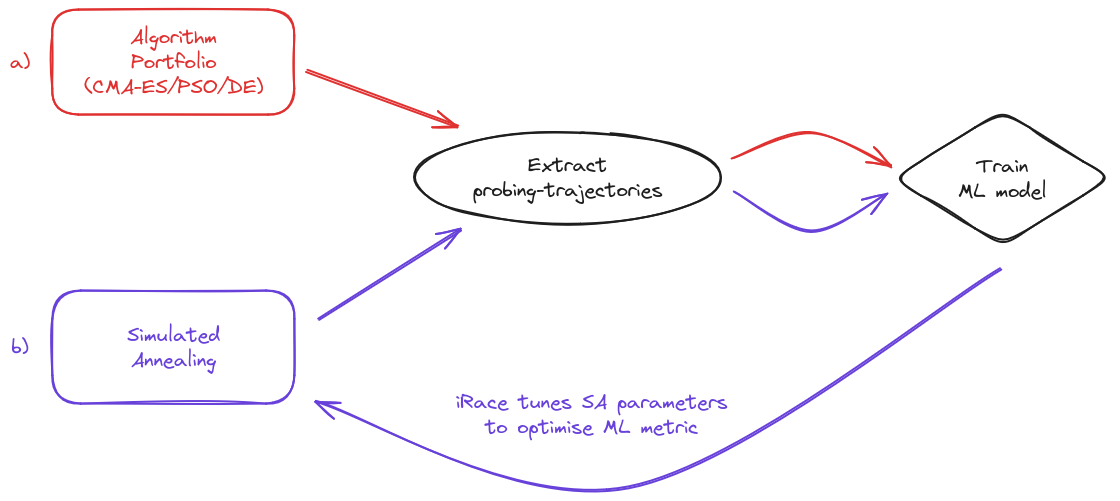}
  \caption{Two approaches to training an algorithm selector are compared. In (a), short trajectories are extracted by running solvers from the portfolio of interest which are then used to train a classifier to predict the best solver. In (b), trajectories are produced from running a simulated annealing algorithm whose parameters are tuned via \textit{irace}~\cite{irace} to optimise a machine-learning metric.
  }
  \label{fig:teaser}
\end{teaserfigure}

\maketitle

\section{Introduction}
\label{sec:intro}
\textit{Algorithm-selection} --- the process of selecting the best algorithm to solve a given problem instance --- is motivated by the fact that algorithms in a portfolio deliver complementary performance on diverse problem instances and was first recognised  by Rice~\cite{Rice76} in 1976. Algorithm-selectors usually take the form of a model that given an input describing an instance,  either predict the label of the best solver (classification) or its performance (regression)~\cite{Tanabe22}.  Typically models are trained using a feature-vector that describes the instance. In continuous domains, Exploratory Landscape Analysis~\cite{mersmann_exploratory_2011} is often used to create features, whereas in combinatorial domains, it is more typical to use hand-designed features that are specific to the domain. 

However, feature-based selection poses a number of issues. Firstly, in some domains, specifying features is not intuitive, and it can be difficult to create a sufficient number to train a model. Conversely, in other domains,  feature-selection methods are required to select the most informative ones~\cite{LoreggiaMSS16,KerschkeKBHT18,smith2014towards}. Secondly, instance features often do not correlate well with algorithm performance data. For instance, Sim {\em et. al.}~\cite{SimH22} demonstrate in the TSP domain that instances that are close in the feature-space can be very distant in the performance space (i.e., the Euclidean distance between their feature-vectors is small while the distance between the performance of two algorithms on the instances is very large). Thirdly, while features can capture nuances of a fitness landscape~\cite{mersmann_exploratory_2011}, they are usually defined \textit{independently} of any solver~\cite{JankovicVKNED22}, which appears problematic if the goal is to inform solver prediction.

To address this, some recent work proposes to use information extracted from fitness trajectories obtained by running a solver for a short period of time to train selectors~\cite{JankovicVKNED22,KostovskaJVNWED22,RenauH24Arxiv}: these methods directly capture something about the search process of a solver on an instance.  Time-series features can be extracted from the trajectories and used to train a selector~\cite{Nobel0B21,CenikjPDKE23,RenauH24Arxiv} or alternatively, the trajectory can be used directly~\cite{RenauH24Arxiv}, avoiding the need for feature-calculation. However, for trajectory-based data to be useful to train a model, it must be \textit{discriminatory} --- despite the success of some of the approaches just mentioned above, it is clear that many meta-heuristic solvers might produce similar trajectories on an instance, particularly given their stochastic nature. 

In this article, we propose a method that harnesses the benefits that trajectory-based approaches can bring (i.e. that they reflect algorithm behaviour) but also addresses the weaknesses just raised. Our method is inspired by research in other fields where \textit{meta-algorithms} have been used to learn a transformation that enables an existing algorithm or model to improve its performance. For example, in the domain of classification, ~\cite{lensen2019can} use genetic programming to learn a low-dimensional embedding of ML datasets such that when a classifier is applied in the learned space, classification metrics are improved. In robotics, ~\cite{bossens2020learning,bossens2022quality} describe a meta-approach to solving control problems in which an EA learns a definition of an archive such that when a MAP-Elites algorithm~\cite{mouret2015illuminating} is applied in the defined archive, objective performance is improved. Also in robotics, Meyerson \textit{et. al.}~\cite{Meyerson16}  propose a method for meta-learning behaviour descriptors which are then used by novelty-search  algorithm~\cite{Lehman2011} to solve a maze-navigation task. As described in Figure~\ref{fig:teaser}, our proposed method uses the well-known algorithm-configuration tool irace~\cite{irace} to tune the parameters of a simple Simulated Annealing (SA) algorithm~\cite{KirkpatrickGV83} such that it produces \textit{short} trajectories that when used to train an algorithm-selector or regressor, optimise the model metrics (e.g. regression or classification). The goal of tuning the SA algorithm is therefore to produce short, discriminatory trajectories that can be cheaply calculated. Following training, when new test instances arrive, the tuned SA algorithm is run to create an input trajectory to a model which is trained to either predict performance or the label of the best solver from a portfolio of interest.

We evaluate the proposed approach on the well-known BBOB test-suite~\cite{bbob-functions} to learn models for both classification and regression. Models are trained either using the trajectory from the tuned SA algorithm directly or by extracting time-series features from the tuned trajectory. As a baseline, we compare the results of the meta-algorithm to models that directly use ELA features and to previous results that use trajectory data obtained by running all solvers in a portfolio.  We show the meta-algorithm outperforms ELA-based models in all experiments; obtains better RMSE than single trajectory-based methods relying on the solver portfolio, and provides  similar or better performance to a multi-trajectory input method but at a considerable less computational cost.

To the best of our knowledge, this is the first time such an approach has been used to improve algorithm-selectors. It addresses several weaknesses in current practice, such as recognising the need to use training data which reflects the behaviour of a solver (i.e. is `algorithm-centric'; the need to have a cheap method of computing input to a model; ensuring the inputs are discriminatory and finally, that 
scales with the size of the portfolio of solvers (unlike the approach described in~\cite{RenauH24Arxiv} which requires trajectories to be computed for every solver in the portfolio to create input to a model).

\paragraph{Reproducibility:} code, data, and additional plots are available at~\cite{dataSA}.

\section{Related Work}
\label{sec:related}

We focus on related work in algorithm-selection in the continuous domain given this is the subject of our study. A more general overview of algorithm-selection (which includes models for both classification and performance prediction) can be found in~\cite{kerschke_automated_2019}.

In continuous optimisation, the vast majority of previous work focuses on the use of \textit{landscape-features} to train selectors.  Exploratory Landscape Analysis (ELA)~\cite{mersmann_exploratory_2011} is a popular method for extracting features, which has grown over the years with a gradual introduction of new features~\cite{munoz_exploratory_2015,kerschke_detecting_2015,DerbelLVAT19}.  The reader is referred to~\cite{renauPhD} for formal definitions of the most used features and their properties. The use of ELA features has been shown to be successful in both algorithm configuration~\cite{belkhir_per_2017} and algorithm selection~\cite{kerschke_automated_2019} on benchmark data as well as real-world optimisation problems~\cite{RenauDPSDD22}.
The main drawback of ELA however is the overhead cost induced by the feature computation: typically this involves sampling a large number of points in order to compute features then discarding the samples, which can be wasteful. In combinatorial optimisation, feature-free approaches to algorithm-selection are beginning to emerge~\cite{alissa2019algorithm,seiler2020deep, zhao2021towards,AlissaSH23}  that use deep-learning (DL) models but have the disadvantage that deep-learning models can be very difficult to explain.

All of the approaches just described take an instance-centric view: that is, the input to a selector is independent of the execution of any algorithm. As already noted in~\ref{sec:intro},  this can be problematic, given that previous work has suggested that there is not necessarily a strong correlation between  the distance of two instances in a feature-space and the distance in the performance space according to a chosen portfolio of solvers~\cite{SimH22}. 

Some recent work has begun to address this, using information derived from running a solver as input to a selector. For example, recent work from Jankovic \textit{et. al.}~\cite{JankovicVKNED22,JankovicED21} proposes  extracting ELA features from the search trajectory of an algorithm. In~\cite{JankovicED21},  trajectories are obtained using half of the available budget and combined with the state variables of the algorithm to train a performance predictor.
However, their approach was outperformed by classical ELA features computed on the full search space.
In~\cite{JankovicVKNED22}, ELA features were computed from short trajectories of $30d$ points: this information was then used to train a model that predicted an algorithm to warm-start a search process.
Cenikj {\em et. al.}~\cite{CenikjPDKE23} log algorithm trajectories using a large budget which is used to construct time-series of statistics derived from both the population fitness to 
 train a classifier that predicts which of the $24$ BBOB functions the trajectory belongs to.
Their approach successfully outperforms a model which is trained on ELA features extracted from algorithm trajectories rather than on statistical information. Kostovska {\em et. al.}~\cite{KostovskaJVNWED22} also use time-series features to train a model: the features are extracted from state variables of CMA-ES to perform a per-run algorithm selection with warm-starting, which demonstrates similar performance to ELA features on the per-run algorithm selection task. Most recently, in~\cite{RenauH24Arxiv}, probing-trajectories capturing the best-so-far or best-per-generation fitness over a small number of generations are used directly to train a selector without extracting any features. Despite the promise of these trajectory approaches, there remain some issues. Firstly, there can be wide variance in trajectories obtained from running the same solver on an instance multiple times, hence randomly sampled trajectories might not discriminate between solvers. Secondly, in the approach described in~\cite{RenauH24Arxiv}, models are trained using a concatenated list of short trajectories obtained from each solver in the portfolio as input. Clearly, this does not scale with increasing sizes of portfolios, where the cost of computing even short trajectories over multiple solvers could be prohibitive.

Although we are unaware of other work that attempts to improve discrimination between model inputs depending on solvers, it is worth mentioning research from other domains which uses a meta-algorithm approach in which a process is nested inside an  outer learning loop that learns something that improves the performance of the inner process. As noted in Section~\ref{sec:intro}, these approaches are common both in Evolutionary Robotics~\cite{Meyerson16,bossens2020learning,bossens2022quality} and particularly in the Quality-Diversity literature~\cite{pugh2016quality}. The latter requires a space to be defined in which an algorithm searches for diverse but high-quality solutions. A meta-algorithm searches for the most appropriate space in which to run the optimiser of interest.  Our work builds on this idea by using a meta-algorithm to tune a simulated annealing algorithm to produce trajectories which are discriminatory with respect to solvers in a portfolio of interest and therefore enable better algorithm-selection models to be trained.

\section{Methods}
In this section, we describe the methods used to generate data, the portfolio of solvers of interest and the machine-learning models used.

\subsection{Searching for Discriminatory Trajectories}
\label{sec:tuning}

As shown in Figure~\ref{fig:teaser}, the meta-approach we propose tunes a Simulated Annealing~\cite{KirkpatrickGV83} algorithm to produce discriminating trajectories. We chose SA both for its simplicity and for the ability to easily tune its hyper-parameters. To minimise the computational effort required to compute trajectories which are used as input to a model but are not used to solve an instance, we minimise the length of the trajectories optimised by SA. 
We use the dual annealing algorithm from scipy~\cite{scipy}~\footnote{version 1.10.1} to provide the trajectories. The search over the configurations of the dual annealing algorithm is performed using the default irace~\cite{irace} package~\footnote{version 3.5}, setting the maximum number of experiments to $5{,}000$.
The parameters involved in creating differentiating trajectories and their possible values are:
\begin{itemize}
    \item The \emph{length of the trajectory} $n \in [5,100]$. The length is restricted to $100$ to provide short trajectories with a low budget of function evaluations.
    \item The SA \emph{initial temperature} $T \in [0.02, 5e4]$.
    \item The SA visiting distribution parameter $v \in [1.5,2.5]$.
    \item The SA acceptance parameter $a \in [-1.1e4, -5]$.
\end{itemize}

We extract two kinds of trajectories: the `best' trajectory records the best fitness seen so far on an instance during a run of the SA algorithm, while the `current' trajectory records the current fitness value (given that SA can accept worse solutions).  The fitness function used in \textit{irace} is the classification accuracy of a model trained on a set of trajectories for a model that predicts the label of the best solver, and the RMSE in a performance prediction task. The  method for determining the labels/performance data is described in the next section.

\subsection{Data}
\label{sec:data}

We consider the first $5$ instances of the $24$ noiseless Black-Box Optimisation Benchmark (BBOB) functions in dimension $d=10$ from the COCO platform~\cite{cocoJournal} as a test-bed.
For each instance, we collect data from running three algorithms: CMA-ES~\cite{HansenO01}, Particle Swarm Optimisation (PSO)~\cite{KennedyPSO95}, and Differential Evolution (DE)~\cite{StornP97}. Each algorithm is run $5$ times per instance.
Following the method described in~\cite{RenauH24Arxiv}, we record the best median target value obtained from running CMA-ES, PSO and DE after $100{,}000$ function evaluations. Note that although this is a very large number of evaluations, its only purpose is to provide a `true' label or performance value for each algorithm to train and evaluate models. The data obtained is used to label each instance with the best-performing algorithm to train a classification model to predict the best solver and as the performance value for a prediction task.

We compare the results of our meta-trajectories approach to training models that  use (1) probing-trajectories directly to train models, (2) time-series features extracted from the trajectories and (3) state-of-the-art ELA feature-vectors as input.

We obtain trajectory data directly from~\cite{dataDiederick} which records search-trajectories per run. Note that some automated algorithm configuration was performed by~\cite{dataDiederick} before they collected this data, see~\cite{Vermetten2022} for further details. ELA feature-data for the BBOB suite is extracted from~\cite{dataEvoStar}.
For each feature, $100$ independent values are available per function instance which were sampled using the \sobol low-discrepancy sequence. We use data from the $10$-dimensional functions with a feature computation budget of $30\times d$ and the general recommendation for feature computation $50\times d$~\cite{kerschke_low-budget_2016}.
We select $10$ cheap features based on their expressiveness and invariance to transformation, described in~\cite{RenauDDD21}. 

\section{Machine Learning Tasks}
\label{sec:task}

We use the trajectories obtained from the tuned SA algorithm to train two types of ML models: (a) a classifier that predicts the best solver and (b) a regressor that predicts performance.  The following section describes the model inputs, the models used,  and the validation procedures that are applied.

\begin{table}[]
\caption{Function evaluations budgets for each algorithm in our portfolio for $2$ and $7$ generations.}
\label{tab:algorithms-budget}
\begin{tabular}{|l|l|l|}
\hline
\multicolumn{1}{|c|}{\begin{tabular}[c]{@{}c@{}}Algorithm\\ Trajectory\end{tabular}} & $2$ generations & $7$ generations \\ \hline
CMA-ES                                                                               & $20$              & $70$              \\ \hline
DE                                                                                   & $60$              & $210$             \\ \hline
PSO                                                                                  & $80$              & $280$             \\ \hline
ALL                                                                                  & $160$            & $560$             \\ \hline
\end{tabular}
\end{table}

\subsection{Inputs for ML models}
\label{sec:input}
Models can be trained with three types of input:

\begin{itemize}
\item \textit{Raw Probing-Trajectories}:  these consist of a time-series of fitness values from the first $n$ function evaluations of an algorithm. Two types of trajectory are recorded.  The `\textit{current}' trajectory simply logs the fitness obtained at every function evaluation in the order that they occur. On the other hand, the `best' trajectory records the \textit{best-so-far} fitness seen during a run after each function evaluation.  We use the labels `best' and `current' to refer to these trajectories from this point onwards.

These trajectories are obtained from the SA algorithm or directly from the data published by~\cite{dataDiederick}.
This archive contains data describing individual trajectories of CMA-ES, PSO, and DE.
Following the approach of~\cite{RenauH24Arxiv}, we also consider `concatenated' trajectories from each of three algorithms which we refer to as the `ALL' trajectory (i.e. it combines individual trajectories into a single new trajectory that can be used to train a model).
For the portfolio solvers CMA-ES/PSO/DE,  we use trajectories obtained over $2$ generations and separately over $7$ generations.
The data we obtained from~\cite{dataDiederick} was generated using a different population size for each solver (due to parameter tuning) hence the length of the trajectories generated for each solver are slightly different. The values are provided in Table~\ref{tab:algorithms-budget}.

The input to a model is always a time-series representing one algorithm run. As just described, the length of the time-series depends on the number of generations and population size used in the algorithms.

\item \textit{Time-series Features}:
\label{sec:ts_features}
As in~\cite{Nobel0B21,RenauH24Arxiv}, we also  extract time-series \textit{features} from all trajectories defined in the previous section, using the \emph{tsfresh} Python package~\footnote{version 0.20.1}, and perform feature selection using the \emph{Boruta} Python package~\footnote{version 0.3}. Input is a vector of features, representing one algorithm run.

\item \textit{ELA Features}
\label{sec:ela}
From the $100$ $10-$dimensional feature vectors that are available for each BBOB instances (see Section~\ref{sec:data}), we randomly sample $5$ vectors.
This ensures fair comparison between features and trajectories as data from $5$ runs is available for calculating trajectories. We compare two budgets for features: the recommended budget of $50d = 500$ sample points and $30d = 300$ sample points.  In this case, the  input is a $10-$dimensional vector of features, representing one function instance.
\end{itemize}

\subsection{Models}
\label{sec:classif}
The \textit{classification } task is a typical algorithm-selection task, i.e., given an input, output the best algorithm to use from a given portfolio. As noted previously, the best algorithm is defined as the algorithm having the lowest median performance after $100{,}000$ function evaluations. There is no under-represented label: out of the $24$ functions, CMA-ES is the best performing algorithm for $11$ functions, DE for $7$, and PSO for $6$. We use the \textit{classification accuracy} as the metric to evaluate the model. The model used depends on the type of input:

\begin{itemize}
\item {\textit{Feature-based}}: inputs can be derived from calculating either ELA features or time-series features extracted from the probing-trajectories.
We train a default Random Forests~\cite{Breiman01} from the \emph{scikit-learn} package~\cite{scikit-learn}\footnote{version 1.1.3} as performed in~\cite{BelkhirDSS17,RenauDPSDD22}.
We train separate models using ELA features or time-series features.

\item \textit{Time-series-based}: for trajectory-based input, we use a specialised time-series classifier, specifically the default Rotation Forests~\cite{RodriguezKA06} from the \emph{sktime} package\footnote{version 0.16.1}. Although other time-series classifiers could have been chosen we selected this as it is closely related to the classifier used for features.
\end{itemize}

The \textit{regression} task consists of a performance prediction of one algorithm from our portfolio after $100{,}000$ function evaluations. Here, the performance measure used for the regression task is the \emph{Root Mean Squared Error (RMSE)}.
Note the performances of algorithms are not normalised.

\begin{itemize}
\item\textit{Feature-based}: we use the default Random Forests~\cite{Breiman01} from the \emph{scikit-learn} package~\cite{scikit-learn}, following the method used in in~\cite{JankovicED21, renauPhD}. Models were trained using ELA features and time-series features. As the latter did not lead to any improvements on the RMSE, results are not shown for this model for space reasons.

\item\textit{Time-series-based}: we use a specialised time-series regressor, specifically the Time Series Forests from the \emph{sktime} package\footnote{version 0.16.1}.
As for the classification, it is chosen for its closeness to the classifier used for features.
\end{itemize}

\subsection{Validation}
Both classification and regression share the same validation procedure.
Following the method described in~\cite{KostovskaJVNWED22}, we perform a \emph{leave-one-instance-out (LOIO) cross-validation}. All Classifiers and regressors are trained using data from all runs of the $24$ functions on all except one instance. The runs from the $24$ functions on the instance left out are used as the validation set.
Overall, $24\times (5-1)\times 5 = 480$ inputs are used to train the model while the remaining $ 24\times 1\times 5 = 120$ inputs are used for validation.

\section{Results}
\label{sec:results}

In this section, we present the results obtained for both ML tasks, i.e., algorithm selection performed with a classifier (Section~\ref{sec:select}) and performance prediction of the three algorithms in our portfolio (Section~\ref{sec:regr}).

\subsection{Tuned SA Parameters}
\label{sec:param}

The motivation for the meta-algorithm proposed is to learn hyper-parameter values for an SA algorithm that produces discriminatory trajectories. Using \textit{irace}, we tune SA separately on eight different tasks: classification using best/current trajectories; regression for each of three solvers, using best/current trajectories).

\begin{figure}
\centering
\begin{subfigure}{.33\textwidth}
  \centering
  \includegraphics[width=\linewidth]{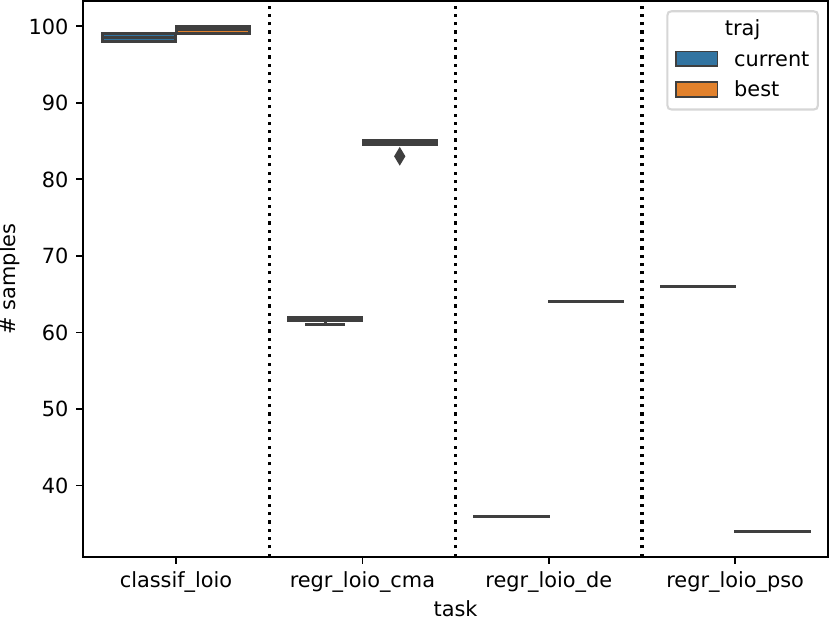}
  \caption{Number of samples}
  \label{fig:samples}
\end{subfigure}
\begin{subfigure}{.33\textwidth}
  \centering
  \includegraphics[width=\linewidth]{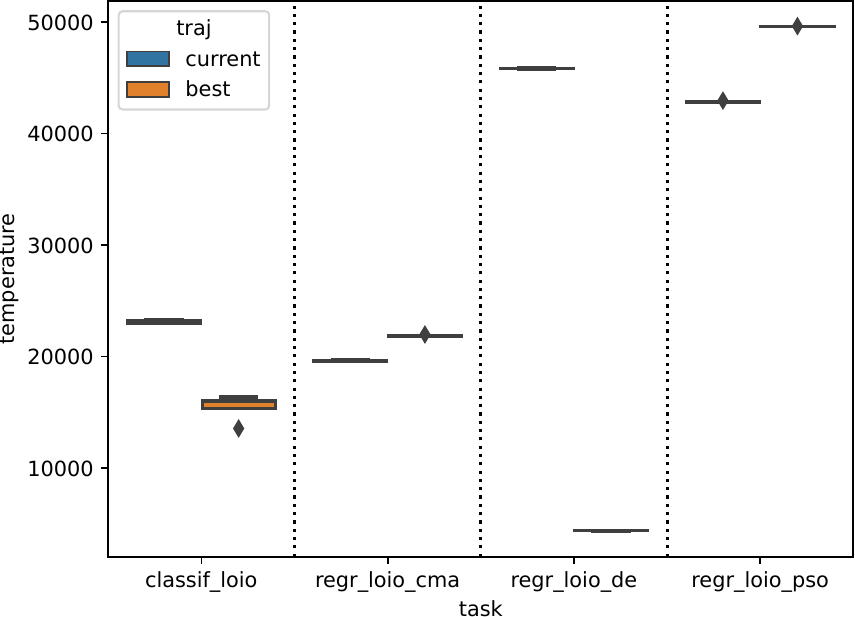}
  \caption{Temperature}
  \label{fig:temp}
\end{subfigure}
\begin{subfigure}{.33\textwidth}
  \centering
  \includegraphics[width=\linewidth]{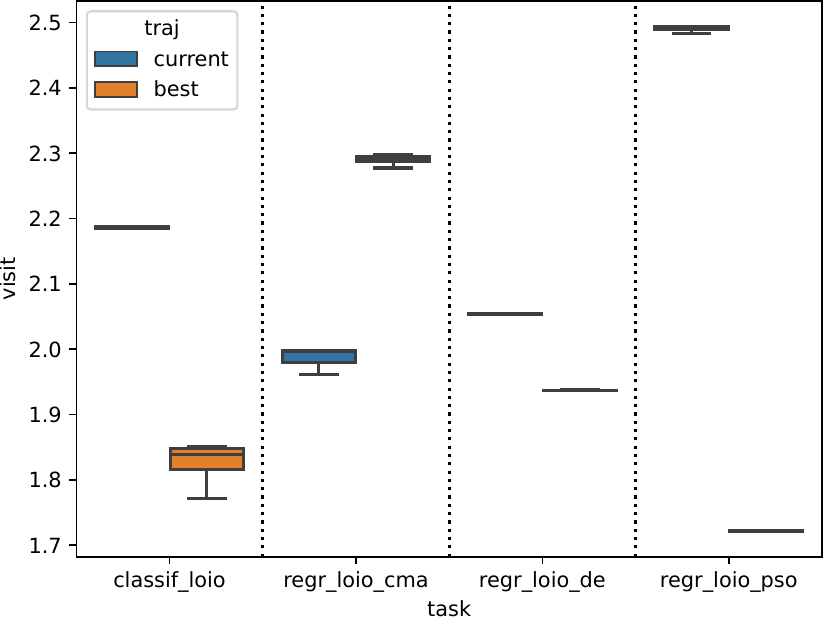}
  \caption{Visit}
  \label{fig:visit}
 \end{subfigure}
 \begin{subfigure}{.33\textwidth}
  \centering
  \includegraphics[width=\linewidth]{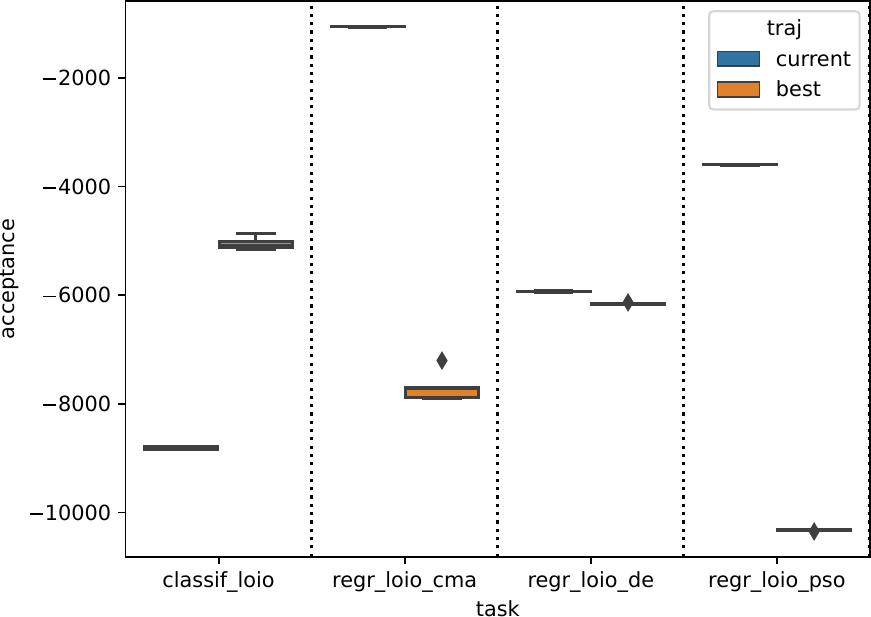}
  \caption{Acceptance}
  \label{fig:accept}
 \end{subfigure}
 \caption{Distributions of SA parameters found by \textit{irace} for the different ML tasks: classification, performance prediction of CMA-ES, PSO, and DE.}
 \label{fig:sa_param}
\end{figure}

The parameters obtained in each case are shown in Table~\ref{tab:SA_params} and used to create test trajectories for all results shown below. It is clear from this table that for each scenario, a unique set of hyper-parameters is obtained.  By default, \textit{irace} outputs the four best configurations found. However in two tasks, \textit{irace} returns fewer configurations: three for the prediction of CMA-ES performances using the `current' trajectory and two for the prediction of DE performances using the `current' trajectory.
Figure~\ref{fig:sa_param} shows the distributions of the four tuned parameters (i.e., number of samples, temperature, visit, and acceptance) for each task across all output configurations.
We observe that the tuned parameters vary widely with respect to the default configuration of SA that uses $100$ samples, a temperature of $5{,}230$, a visit of $2.62$ and an acceptance of $-5$.  The variance of the distribution of configurations found by \textit{irace} is small for each task indicating that there do not seem to be multiple optima in the search-space.

\begin{table}[]
\caption{SA parameters for the best configuration found by \textit{irace} for each ML task.}
\label{tab:SA_params}
\begin{tabular}{|l|l|l|l|l|}
\hline
Task                                                                & \# samples & T & visit  & acceptance  \\ \hline
Default                                                             & $100$        & $5230$        & $2.62$   & $-5$          \\ \hline
\begin{tabular}[c]{@{}l@{}}Classification\\ current\end{tabular}    & $99$        & $23{,}318.74$  & $2.185$ & $-8{,}848.35$  \\ \hline
\begin{tabular}[c]{@{}l@{}}Classification\\ best\end{tabular}       & $99$        & $15{,}912.03$  & $1.831$ & $-5{,}110.81$  \\ \hline
\begin{tabular}[c]{@{}l@{}}Regression\\ CMA-ES current\end{tabular} & $62$        & $19{,}578.93$  & $1.998$ & $-1{,}053.54$  \\ \hline
\begin{tabular}[c]{@{}l@{}}Regression\\ CMA-ES best\end{tabular}    & $85$        & $21{,}838.23$  & $2.299$ & $-7{,}878.54$  \\ \hline
\begin{tabular}[c]{@{}l@{}}Regression\\ PSO current\end{tabular}    & $66$        & $42{,}832.28$  & $2.492$ & $-3{,}601.99$  \\ \hline
\begin{tabular}[c]{@{}l@{}}Regression\\ PSO best\end{tabular}       & $34$        & $49{,}602.83$  & $1.722$ & $-10{,}325.21$ \\ \hline
\begin{tabular}[c]{@{}l@{}}Regression\\ DE current\end{tabular}     & $36$        & $45{,}778.60$  & $2.053$ & $-5{,}925.62$  \\ \hline
\begin{tabular}[c]{@{}l@{}}Regression\\ DE best\end{tabular}        & $64$        & $4{,}315.52$  & $1.938$ & $-6{,}135.95$  \\ \hline
\end{tabular}
\end{table}

\subsection{Algorithm Selection}
\label{sec:select}

\begin{figure}
\centering
\begin{subfigure}{.45\textwidth}
  \centering
  \includegraphics[width=\linewidth]{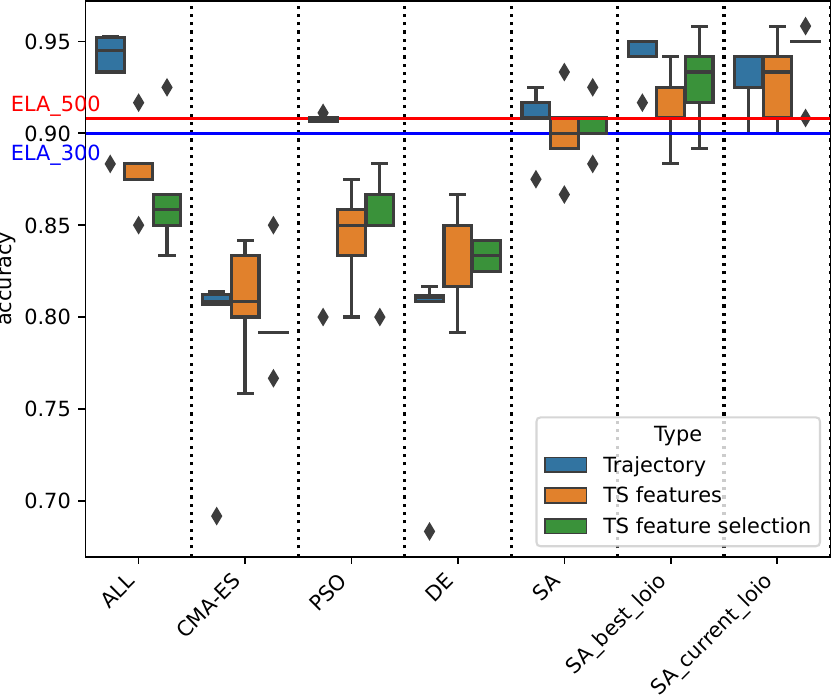}
  \caption{Best trajectory for $2$ generations}
  \label{fig:select_2}
\end{subfigure}
\begin{subfigure}{.45\textwidth}
  \centering
  \includegraphics[width=\linewidth]{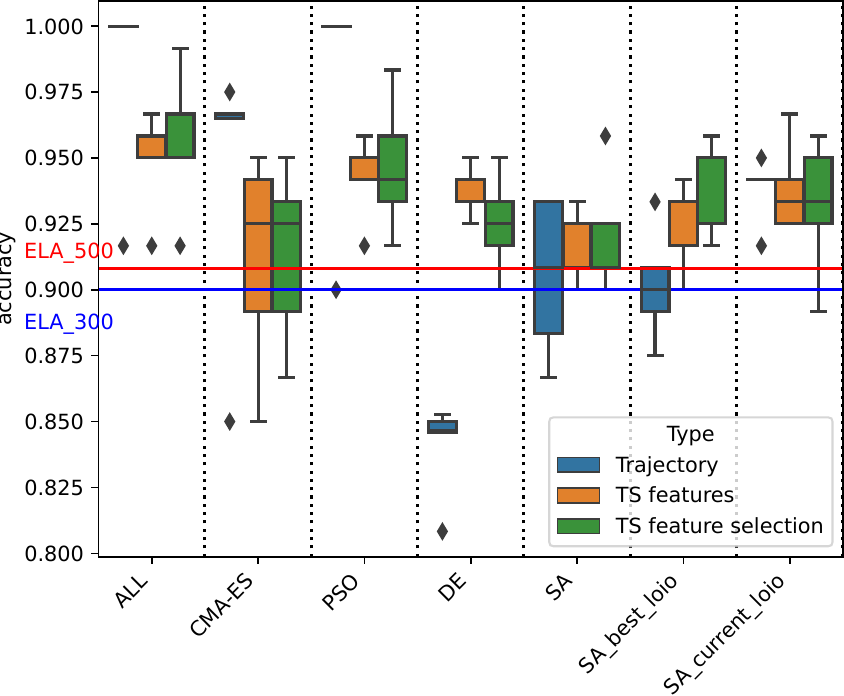}
  \caption{Current trajectory for $7$ generations}
  \label{fig:select_7}
 \end{subfigure}
 \caption{Accuracy of classification on the LOIO cross-validation for best-so-far and current probing-trajectories, time series features and time series feature selection for $2$ and $7$ generations. Median ELA feature accuracy is represented by lines for $300$ and $500$ function evaluations.}
 \label{fig:select}
\end{figure}

Figure~\ref{fig:select} compares the classification accuracies obtained with ELA features and trajectories (individual trajectories from algorithms in the portfolio, concatenated trajectories from algorithms in the portfolio, default SA trajectories used as baseline ($SA$), and tuned SA trajectories ($SA-best, SA-current$).
Following~\cite{RenauH24Arxiv}, we display results for trajectories obtained over $2$ and $7$ generations for each of the portfolio algorithms. `Best' trajectories are displayed for $2$ generations (Figure~\ref{fig:select_2}) and `current' trajectories for $7$ generations (Figure~\ref{fig:select_7}) to save space (additional plots are available at~\cite{dataSA}).  
For each of the different methods of obtaining trajectories, we show boxplots for classifiers trained on trajectories and time-series features (with/without selection), and indicate median ELA performance as a line.

Figure~\ref{fig:select_2}  where trajectories are obtained from only $2$ generations shows that classifiers trained on SA trajectories (tuned or by default) and on the \textit{`ALL'} trajectories outperform classifiers trained on ELA features at both budgets with respect to  median fitness.  Recall however that the SA trajectories are obtained from only $100$ or fewer samples depending on the tuned sample-size, i.e. they use three to five times less budget than ELA features.
$SA\_best\_loio$ has a median accuracy of $95\%$ compared to  $94.5\%$ for the `ALL' classifier. $SA\_current\_loio$ has a $92.5\%$ median accuracy using the raw-trajectory,  rising to $95\%$ accuracy when time-series feature selection is used.
The tuned SA trajectories result in better accuracy than the untuned version (as expected). Classifiers trained on the raw trajectory data outperform the feature-based approaches as a rule of thumb. Training on individual trajectories performs poorly in comparison in each case.

Figure~\ref{fig:select_7} shows results when trajectories are collected over $7$ generations using `current' trajectories.  In contrast to the previous results, all methods except the DE classifier have a median accuracy better than those obtained using ELA features. While the SA trajectories outperform ELA features, they are outperformed by other classifiers:  $SA\_current\_loio$ has a median accuracy of $94.2\%$ which is lower than all other classifiers except for the one trained on DE trajectories.  Nevertheless, we highlight that the SA methods require significantly fewer function evaluations to achieve this result than every other algorithm presented here except CMA-ES.
SA is also more robust than CMA-ES which has one outlier with $85\%$ accuracy, while SA consistently performs above $90\%$. 

Overall, tuning SA to find discriminating  trajectories to train an algorithm-selector is a good low-budget approach  in that it requires significantly fewer function evaluations than ELA features and some of the trajectories approaches. It also provides some confidence the trajectories are discriminatory: this is not guaranteed when calculating trajectories from the three solvers in the portfolio due to the stochastic nature of the methods. Furthermore, if the portfolio of algorithms is large, this introduces an additional burden due to the number of combinations of trajectories that could be concatenated, making it more difficult to select the best combination to train with. The budget required for this may be more expensive than calculating  ELA features for a large portfolio.

\subsection{Performance Prediction}
\label{sec:regr}

\begin{figure}
\centering
\begin{subfigure}{.4\textwidth}
  \centering
  \includegraphics[width=\linewidth]{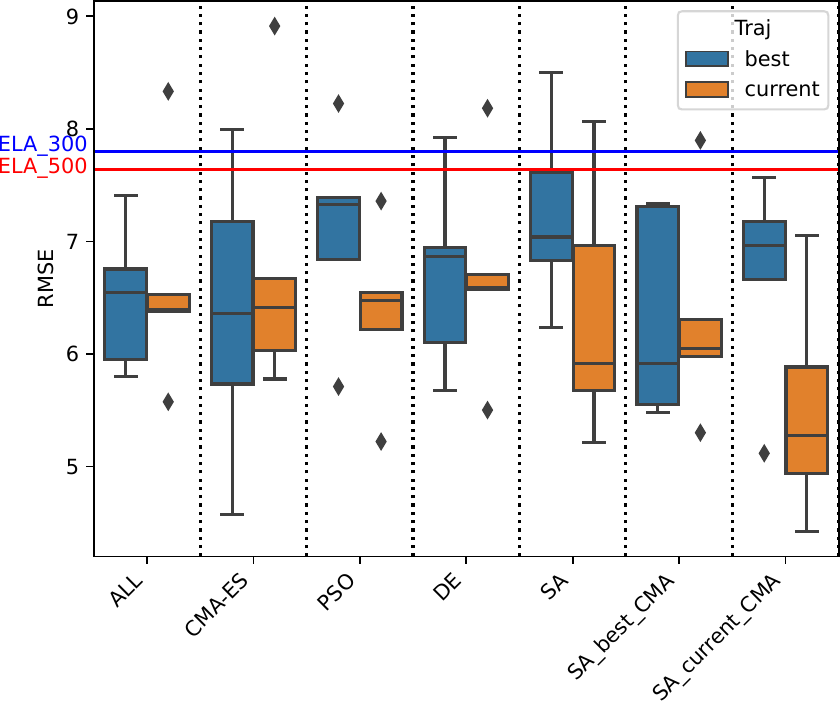}
  \caption{CMA-ES}
  \label{fig:regr_cma}
\end{subfigure}
\begin{subfigure}{.4\textwidth}
  \centering
  \includegraphics[width=\linewidth]{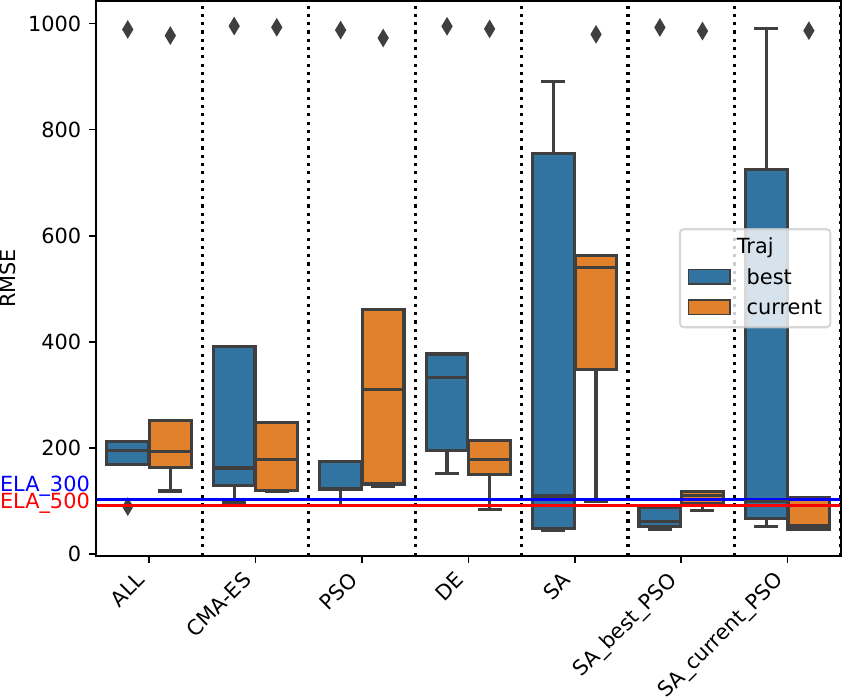}
  \caption{PSO}
  \label{fig:regr_pso}
 \end{subfigure}
 \begin{subfigure}{.4\textwidth}
  \centering
  \includegraphics[width=\linewidth]{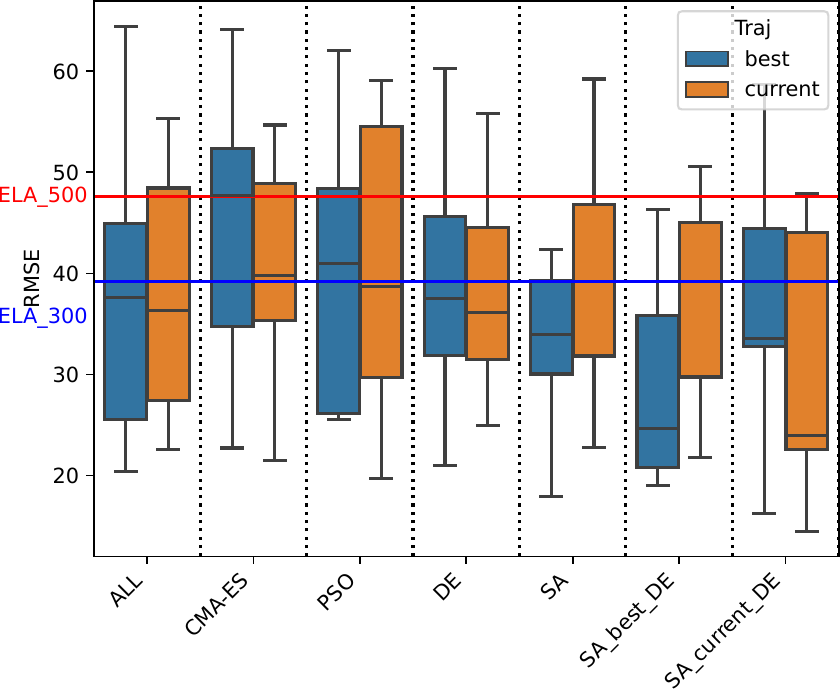}
  \caption{DE}
  \label{fig:regr_de}
 \end{subfigure}
 \caption{RMSE of performance prediction of CMA-ES, PSO, and DE on the LOIO cross-validation for best-so-far and current probing-trajectories for $7$ generations. Median ELA feature RMSE is represented by lines for $300$ and $500$ function evaluations.}
 \label{fig:regr}
\end{figure}

Figure~\ref{fig:regr} shows the RMSE of the performance prediction of CMA-ES, PSO, and DE using trajectories and ELA features for $7$ generations on both `best' and `current' trajectories. Note that this should be minimised. For each solver in the portfolio, we show the RMSE of the predicted performance obtained from the set of regressors\footnote{Similar results are obtained using $2$ generations  --- results are not included for space reasons}. For each regressor, we omit the results from training using time-series features as they do not improve results.

We observe that for CMA-ES (Figure~\ref{fig:regr_cma}) and DE (Figure~\ref{fig:regr_de}) performance predictions, all of the trajectory-based regressors outperform or match ELA performances. However, for PSO, only the tuned SA-based regressors outperform ELA. All of the other methods result in an RMSE that is worse than predicted by the ELA regressors.

The default SA configuration `current' trajectory outperforms all regressors trained on individual trajectories from the original portfolio (CMA-ES/DE/PSO/ALL) with a median RMSE of $5.91$. However, the best result is obtained from the tuned SA approach $SA\_current\_CMA$ with a median RMSE of $5.28$.
Again, we remind the reader that this is also the approach using the smallest budget of function evaluations with $62$ (close to the CMA-ES trajectory $70$ evaluations and nine times less than the `ALL' trajectory).

For predicting DE, the default configuration of SA with the `best' trajectory also outperforms other trajectory-based regressors  where the algorithm comes from the initial portfolio and also outperforms the regressor trained on ELA features  (Figure~\ref{fig:regr_de}).
For this task, both specifically tuned SA methods have similar performance with a median RMSE of $23.95$ for $SA\_current\_DE$ on the `current' trajectory and $24.6$ for $SA\_best\_DE$ on the `best' trajectory.
For this task, it is more advantageous to use $SA\_current\_DE$: its performance is slightly better but its computational cost is lower, i.e., $36$ function evaluations against $64$ for $SA\_best\_DE$.
Using only $36$ function evaluations (i.e. the tuned value from \textit{irace}), $SA\_current\_DE$ uses almost half CMA-ES budget to compute trajectories and $15$ times less evaluations than the `ALL' trajectory.

Prediction of performances of PSO (Figure~\ref{fig:regr_pso}) is the only regression task where only the specifically tuned SA methods outperform ELA features. All other models trained using trajectories obtain worse RMSE.
Both tuned SA methods achieve similar RMSE with a median of $54.4$ for $SA\_current\_PSO$ and $60.6$ for $SA\_best\_PSO$.
We suggest $SA\_best\_PSO$ seems more beneficial from a computational point of view as it requires only $34$ function evaluations against $66$ for $SA\_current\_PSO$.

To summarise, it is clear that trajectory-based algorithm selection and trajectory-based performance prediction provide performances that are equal to or better than the metrics obtained using ELA features as input, using far less computational budget.
Tuning SA to obtain discriminating trajectories is the only trajectory-based approach to consistently outperform ELA features, and also outperforms other trajectory-based models.

\subsection{Transfer Learning to reduce computation}
\label{sec:cross}
In previous sections, we showed that tuning SA to produce discriminatory trajectories outperforms the state-of-the-art. For the regression task, SA was tuned separately to create trajectories to train regressors for each of the three solvers (CMA-ES, DE, PSO).  The ML literature suggests that often there is potential for transfer-learning (TL)~\cite{niu2020decade} between models ---  this aims to improve the performance of learners on target domains by transferring the knowledge contained in different but related source domains. Therefore
we now assess whether TL can be exploited to reduce the computational burden of tuning per solver. For example, we examine whether the hyper-parameters used to produce discriminatory trajectories for PSO and DE can also be used to predict CMA-ES performances.

\begin{figure}
\centering
\includegraphics[width=\linewidth]{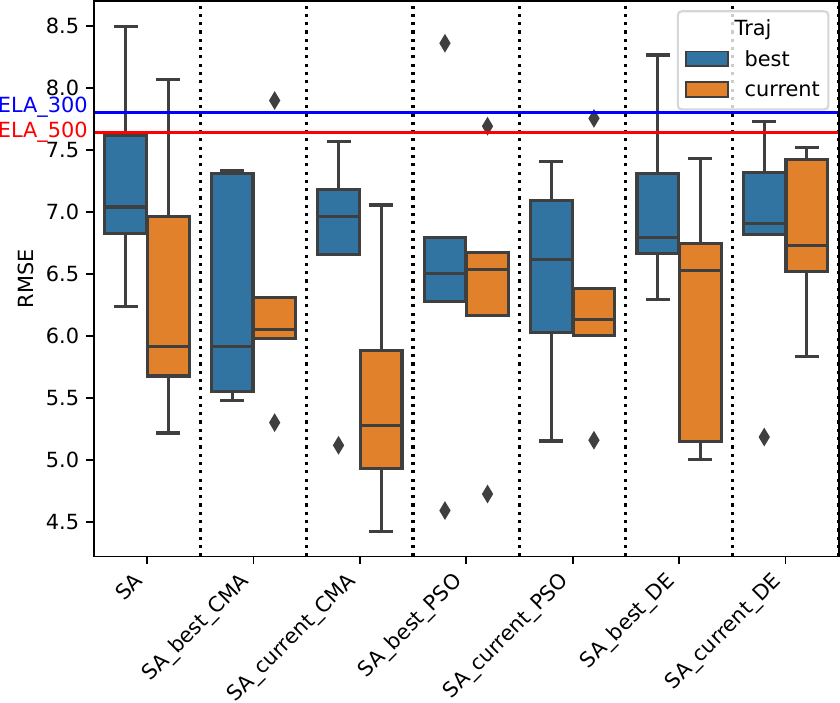}
 \caption{RMSE of performance prediction of CMA-ES on the LOIO cross-validation for best-so-far and current probing-trajectories for all tuned SA (including other tasks). Median ELA features RMSE are represented by lines for $300$ and $500$ function evaluations.}
 \label{fig:cross_regr}
\end{figure}

Figure~\ref{fig:cross_regr} shows the RMSE for performance prediction of \textbf{CMA-ES} using regressors trained on trajectories tuned on DE, PSO and CMA-ES respectively.

As  expected, the specialists $SA\_current\_CMA$ and $SA\_best\_CMA$  outperform all other configurations (median performance $5.91$, $5.28$ respectively for the `best'  and `current' trajectory models). Nevertheless,  using trajectories obtained from SA with hyper-parameters tuned on DE or PSO to predict CMA-ES performance is surprisingly robust:  using parameters tuned for PSO gives a median RSME of $6.5$ (compared to $5.91$ from the specialist), while parameters tuned for DE have median RMSE $6.79$. We also note that tuning on a 'best' trajectory and re-using the parameters to generate a `current' trajectory for a solver often results in a noticeable loss in RSME: for example, using parameters tuned on the CMA-ES `current' trajectories to obtain trajectories for CMA-ES `best' increases the median RMSE from $5.91$ to $6.96$.
Finally, it is worth observing that \textit{all} methods provide better results than regressors trained on ELA features: that is, tuning SA to generate discriminatory trajectories for \textit{any} of the solvers in the portfolio provides a set of hyper-parameters that can be re-used to create trajectories for other solvers in the portfolio that can be used to predict performance more accurately than ELA features.

Given that the `best' and `current' trajectories have different configurations of SA that can be very different, it seems not possible to transfer the tuned parameters to the other type of trajectory without having a loss in performance.

\section{Conclusion}
\label{sec:conclusion}

We addressed the issue of algorithm-selection and performance prediction in a continuous optimisation setting. 
Previous work~\cite{RenauH24Arxiv} has shown that using data that is algorithm-centric (i.e. trajectories) to train models can outperform ML models trained using features (e.g., ELA), with the additional advantage of being low-budget in terms of the number of function evaluations needed to create a trajectory. However there are some weaknesses with this approach: (1) it is difficult to ensure that trajectories are sufficiently discriminatory to train a high-performing model and (2) as a trajectory needs to be generated per solver in a portfolio, the approach does not scale well.

To address this, we proposed a meta-algorithm that tunes the hyper-parameters of a simple solver (Simulated Annealing) to generate trajectories that when used as input data to an ML model, improve its performance metric (either classification or regression). This directly addresses both weaknesses identified above. Firstly, the meta-algorithm learns to create discriminatory trajectories. Secondly, rather than generating trajectories from each solver in a portfolio as in~\cite{RenauH24Arxiv}, only one trajectory needs to be generated using the tuned SA algorithm. Furthermore, the SA trajectories only use $100$ or fewer samples and, therefore are obtained at a very low budget.  We highlight the following findings:

\begin{itemize}

\item The ML performance-metrics obtained using models that use trajectories from a tuned SA algorithm as input outperform the same metrics obtained from models that use ELA features as input, using considerably less budget (at minimum, $\approx 1/3$ of the ELA budget).

\item For classification models that predict the label of the best solver, at low-budget ($2$ generations), models using SA trajectories from the tuned SA algorithm as input have similar median accuracy to those obtained using an `ALL' trajectory, but use $\approx 62\%$ of the budget used by `ALL'.
\item For all three regression models that predict performance (CMA-ES, DE, PSO), the best RMSE is obtained by a model that uses an SA trajectory as input.
\item Although using the meta-algorithm to tune each model individually leads to the best results, the loss in RMSE when using a set of hyper-parameters tuned for one solver to obtain trajectories for a different solver is very small, suggesting that budget could be saved with the trade-off of a small reduction in performance.
\end{itemize}

Obvious next steps include testing the approach on a larger portfolio of solvers and on different benchmarks. With respect to the latter, it would be particularly interesting to consider combinatorial optimisation domains here although this may exclude a comparison to ELA-based methods as the use of landscape features is much less common in combinatorial domains.

Given that the  method relies on creating a time-series of data, then a more thorough evaluation of state-of-the-art time-series classifiers (and tuning of their hyper-parameters) is also likely to improve results.
We investigated tuning the hyper-parameters of a Simulated Annealing  algorithm to create the discriminatory trajectories but clearly, other algorithms could replace this. We stress however that the motivation behind this work is to generate data at a low-cost, therefore this might restrict the type of algorithm that could be used. It would also be interesting to
design a specialist algorithm  whose only purpose would be to generate discriminatory trajectories. Finally, further work in understanding the extent to which transfer-learning could be exploited to further reduce computation would be beneficial.

\section*{Acknowledgements}
The authors are supported by funding from EPSRC award number(s): EP/V026534/1

\clearpage

\bibliographystyle{ACM-Reference-Format}
\bibliography{references}

\end{document}